\pdfoutput=1

\documentclass[11pt]{article}

\usepackage[preprint]{acl}

\usepackage{times}
\usepackage{latexsym}

\usepackage[T1]{fontenc}

\usepackage[utf8]{inputenc}

\usepackage{microtype}

\usepackage{inconsolata}

\usepackage{graphicx}
\usepackage{times}
\usepackage{latexsym}
\usepackage{times}
\usepackage{latexsym}
\usepackage{booktabs}
\usepackage{boxedminipage}
\usepackage{amssymb}
\usepackage{amsmath}
\usepackage{todonotes} 
\usepackage{graphicx}

\usepackage{hyperref}
\usepackage{inconsolata}
\usepackage{pifont}

\usepackage{tabu}
\usepackage{fontawesome}
\usepackage{amsmath}
\usepackage{amsfonts}
\usepackage{amssymb}
\usepackage{enumitem}
\usepackage{listings}
\usepackage{xstring}
\usepackage{graphicx}
\usepackage{pbox}
\usepackage{subcaption}
\usepackage{epstopdf}
\usepackage{xstring}
\usepackage{multirow}

\usepackage{soul}
\usepackage{color}
\usepackage{graphicx}
\usepackage{multirow}
\usepackage{multicol}
\usepackage{comment}
\usepackage{todonotes}
\usepackage{listings} 
\usepackage{graphicx}
\usepackage{subcaption}
\usepackage{verbatim}
\usepackage[utf8]{inputenc}

\lstdefinestyle{smalllisting}{
    basicstyle=\small\ttfamily
}

\usepackage{xcolor} 
\definecolor{lightblue}{rgb}{.50,.90,0.51}
\definecolor{tri}{rgb}{.25,.88,.82}
\definecolor{lilac}{rgb}{0.85,0.64,0.85}
\definecolor{atomictangerine}{rgb}{1.0, 0.6, 0.4}

\newcommand{\multihate}{\emph{BanglaMultiHate}}

\lstdefinestyle{pythonstyle}{
  frame=tb,
  language=python,
  aboveskip=3mm,
  belowskip=3mm,
  showstringspaces=false,
  columns=flexible,
  basicstyle={\small\ttfamily},
  numbers=none,
  numberstyle=\tiny\color{gray},
  keywordstyle=\color{dkgreen},
  commentstyle=\color{dkgreen},
  breaklines=true,
  breakatwhitespace=true,
  tabsize=3,
  escapeinside={`}{`},
  breakindent=0pt,
  otherkeywords={these}
}

\title{LLM-Based Multi-Task Bangla Hate Speech Detection: \\
Type, Severity, and Target\\\hphantom{...}
\\\footnotesize{\textcolor{red}{WARNING: This paper contains examples which may be disturbing to the reader}}}

\author{Md Arid Hasan$^1$, Firoj Alam$^2$, Md Fahad Hossain$^3$, Usman Naseem$^4$,\\ 
\textbf{Syed Ishtiaque Ahmed$^1$}\\
$^1$University of Toronto, Canada, $^2$Qatar Computing Research Institute, Qatar \\
$^3$Daffodil International University, Bangladesh $^4$Macquarie University, Australia \\  
{\tt \{arid, ishtiaque\}@cs.toronto.edu, fialam@hbku.edu.qa}\\
}

\begin{document}
\maketitle
\begin{abstract}
Online social media platforms are central to everyday communication and information seeking. While these platforms serve positive purposes, they also provide fertile ground for the spread of hate speech, offensive language, and bullying content targeting individuals, organizations, and communities. Such content undermines safety, participation, and equity online. Reliable detection systems are therefore needed, especially for low-resource languages where moderation tools are limited. In Bangla, prior work has contributed resources and models, but most are single-task (e.g., binary hate/offense) with limited coverage of multi-facet signals (type, severity, target). We address these gaps by introducing the \textit{first multi-task} Bangla hate-speech dataset, \multihate{}, one of the largest manually annotated corpus to date. Building on this resource, we conduct a comprehensive, controlled comparison spanning classical baselines, monolingual pretrained models, and LLMs under zero-shot prompting and LoRA fine-tuning. Our experiments assess LLM adaptability in a low-resource setting and reveal a consistent trend: although LoRA-tuned LLMs are competitive with BanglaBERT, culturally and linguistically grounded pretraining remains critical for robust performance. Together, our dataset and findings establish a stronger benchmark for developing culturally aligned moderation tools in low-resource contexts. For reproducibility, we will release the dataset and all related scripts.

\end{abstract}

\section{Introduction}
\label{sec:introduction}

The rise of social media has increased the spread of harmful online content \cite{walther2022social}, with hate speech emerging as a critical societal issue given its potential to perpetuate discrimination \cite{gelber2021differentiating}, harassment, and violence. Given the large volume of user-generated content, manual moderation is neither scalable nor consistent, highlighting the urgent need for reliable and scalable automated hate speech detection systems. 
Although substantial progress has been achieved in high-resource languages such as English \cite{albladi2025hate}, research effort for low-resource languages like Bangla are relatively limited \cite{sharma2025hate,das-etal-2022-hate-speech,haider-etal-2025-banth,romim-etal-2022-bd}. 

Identifying hate speech in Bangla imposes unique challenges due to its rich morphology, free word order, and code-switching with English and other regional dialects, making it difficult for models trained on other languages to generalize effectively. Furthermore, the scarcity of annotated datasets and the lack of high-quality pretrained resources exacerbate the difficulty of building accurate classification systems \cite{al2024hate}. Existing studies often rely on classical machine learning models \cite{kiela2020hateful, mridha2021lboost, romim-etal-2022-bd}, deep learning models \cite{romim-etal-2022-bd, keya2023gbert}, and adapt pretrained models designed primarily for English \cite{mridha2021lboost}. However, these approaches often fail to capture the cultural, social, and linguistic nuances that shape how hate is expressed in Bangla \cite{al2024hate}, such as context-dependent slurs, metaphorical insults, or region-specific idiomatic usage \cite{jahan-etal-2022-banglahatebert}. Addressing these challenges requires not only improved datasets and resources but also approaches that are sensitive to the sociolinguistic realities of Bangla discourse, ensuring that models move beyond surface-level understanding and engage with the deeper structures of the language.

\begin{figure}[]
    \centering    
    \includegraphics[scale=0.32]{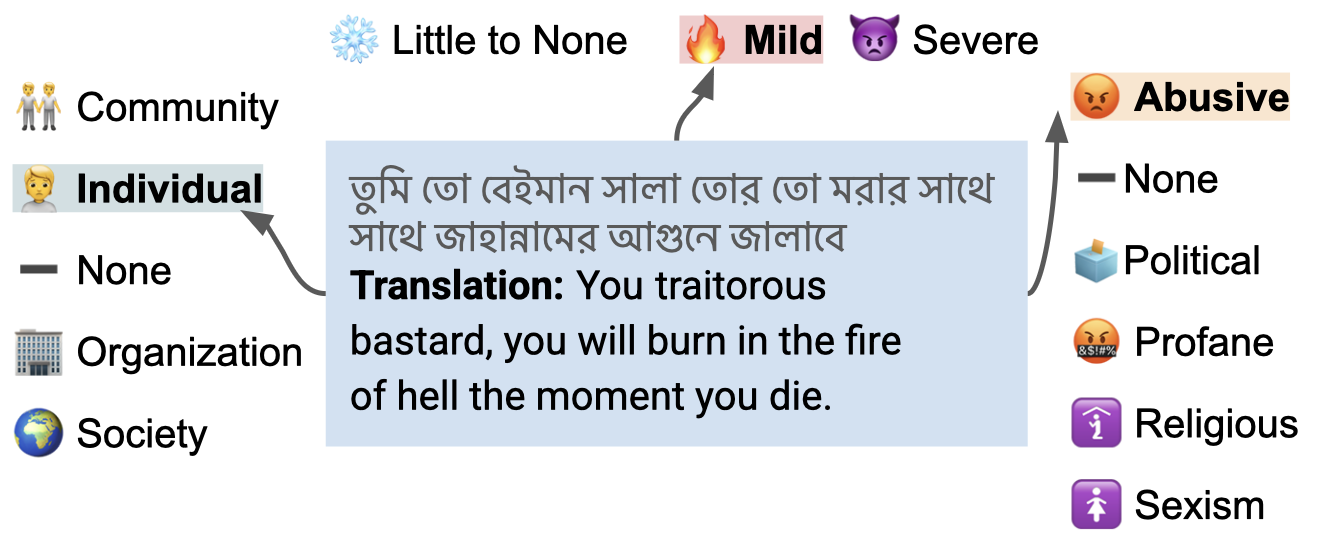}
    \vspace{-0.2cm}
    \caption{An example of hateful comment with its English translation showing type, severity and target of hate.}
    \label{fig:hate_speech_example}
    \vspace{-0.4cm}
\end{figure}

In recent years, the rapid advancement of large language models (LLMs) such as GPT-5, Claude, Gemini \cite{comanici2025gemini}, Llama \cite{dubey2024llama}, and Qwen \cite{yang2025qwen3} has shown remarkable success across a variety of downstream NLP tasks, often demonstrating strong generalization abilities in zero-shot or few-shot scenarios. This has raised important questions about their applicability in sensitive domains such as hate speech detection, particularly for underrepresented languages. However, the zero-shot performance of LLMs in low-resource contexts is often limited and their inability to capture context-dependent information \cite{zahid2025evaluation}. Moreover, hate speech is highly context-dependent and culturally nuanced, making it difficult for LLMs pretrained on high-resource languages to detect, demonstrating the need for targeted adaptation strategies in low-resource and sensitive tasks.

To address these challenges, we developed the first multi-tasks hate speech dataset named \textit{BanglaMultiHate} for Bangla. This dataset is specifically designed to support a variety of classification tasks: \textit{(i)} identifying different types of hate speech, \textit{(ii)} the severity of hate, and \textit{(iii)} determining the target of hate. An example of a hateful comment with type, severity and target is demonstrated is in Figure \ref{fig:hate_speech_example}. 
This study also conducts a comprehensive evaluation of hate speech detection in Bangla across three tasks, utilizing SVM, BanglaBERT, zero-shot settings (Llama3 and Qwen3), and LoRA fine-tuned on these LLMs. Our contributions can be summarized as follows:

\begin{itemize}[noitemsep,topsep=0pt,labelsep=.5em]
    \item We developed the first multi-task hate speech dataset for Bangla and one of the largest manually annotated hate speech datasets, which includes type of hate, severity and target.
    \item We provide comprehensive comparisons of classical, monolingual pretrained, and zero-shot and LoRA fine-tuned approaches using LLMs for Bangla hate speech detection.
    \item We assess the effectiveness of zero-shot inference and LoRA fine-tuning for LLMs, offering insights into their adaptability in low-resource tasks.
    \item We highlight key limitations and trends, demonstrating that while fine-tuned LLMs are comparable to the performance of BanglaBERT, emphasizing the continued importance of culturally and linguistically grounded pretraining for combating online hate speech in low-resource languages.
\end{itemize}

Our findings are summarized as follows:
\begin{itemize}[noitemsep,topsep=0pt,labelsep=.5em]
    \item Fine-tuned monolingual BanglaBERT yields superior performance.
    \item Zero-shot learning failed to perform better than the majority baseline as well as SVM.
    \item SVM performs comparatively better than fine-tuned LLMs on severity and target of hate tasks, while Llama3 performs slightly better on the type of hate task.
    \item Model performance varies significantly with the complexity of the task.
\end{itemize}



\section{Related Work}
\label{sec:related_work}


The identification of offensive language and hate speech has become increasingly important due to the extensive use of social media, which has created an environment in which harmful content can spread rapidly \cite{jiang2024cross}. 
Research on hate speech identification has progressed rapidly over the past decade \cite{fortuna2018survey}, moving from lexicon-based classifiers to transformer models and, more recently LLMs \cite{albladi2025hate}.

\subsection{Models}
Various classical models (such as logistic regression (LR), SVM, and random forest), deep learning models (e.g., LSTM), and transformer-based models (e.g., BERT, XLM-R, MuRIL, AraBERT, etc.) have been studied in the literature. \citet{sharif2021nlp} demonstrated that transformer-based pretrained language models (e.g., Indic-BERT, XLM-R, mBERT) outperform classical models (e.g., LR, SVM). The multi-task learning approach using AraBERT has been studied in Arabic for the identification of offensive language and hate speech \cite{djandji2020multitask}, while random forest, k-nearest neighbors, and MLP classifiers have been studied for offensive language identification from Dravidian code-mixed texts \cite{ssncse2021dravidianlangtech}. \citet{pelicon2021zero} employed mBERT and LASER models for zero-shot cross-lingual transfer learning, demonstrating promising results in languages such as German, Spanish, Indonesian, and Arabic. Similarly, \cite{saumya2021offensive} explores the impact of cross-cultural transfer learning, showing how biases across cultures affect model performance, examining the impact of cross-cultural transfer learning.

\citet{kiela2020hateful} utilizes SVM, CNN, and LSTM models to evaluate performance on hateful content. SVM, naive bayes, and random forest, along with transformation methods have been studied for multi-label hate speech identification \cite{ibrohim-budi-2019-multi}. \citet{mridha2021lboost} employed L-Boost, a modified AdaBoost algorithm combining BERT embeddings with LSTM models, to identify offensive texts in Bangla and Banglish social media content. SVM, LSTM, and Bi-LSTM models have also been analyzed by \citet{romim2020hate} on Bangla YouTube and Facebook comments, with results showing that SVM outperforms LSTM and Bi-LSTM. Furthermore, combining BERT and GRU architectures for hate speech detection has been explored in Bengali social media texts \cite{keya2023gbert}. Explainable hate speech identification has recently attracted attention in the literature \cite{yang2023hare, piot2025towards, sariyanto2025towards}.

\begin{table}[]
\centering
\setlength{\tabcolsep}{3pt}
\scalebox{0.45}{%
\begin{tabular}{lclll}
\toprule
\textbf{Dataset} & \textbf{Size} & \textbf{Type} & \textbf{Labels / Tasks} & \textbf{Source} \\ \hline
BD-SHS \cite{romim-etal-2022-bd} & 50,281 & Comments & \begin{tabular}[c]{@{}l@{}}3-level: HS vs. non-HS; \\ target; HS type\end{tabular} & Social media \\
Bengali Tweets \cite{das-etal-2022-hate-speech} & 10,000 & Code-mixed & Hate/offense detection (binary) & Twitter/X \\
TB-OLID \cite{raihan2023offensive} & 5,000 & \begin{tabular}[c]{@{}l@{}}Transliterated, \\ code-mixed\end{tabular} & \begin{tabular}[c]{@{}l@{}}OLID: Offensive vs. Not; \\ target type (Indiv./ \\ Group/Untargeted)\end{tabular} & Facebook \\
BanTH \cite{haider-etal-2025-banth} & 37,350 & Transliterated & Multi-label target; HS & YouTube \\
BIDWESH \cite{fayaz2025bidwesh} & 9,183 & Dialectal & \begin{tabular}[c]{@{}l@{}}Hate vs. non-hate; \\ $\sim$13 types; 7 targets\end{tabular} & Social media \\ \midrule
BanglaMultiHate (Ours) & 50,746 & Comments & Type, severity, target & YouTube \\ \bottomrule
\end{tabular}
}
\caption{Overview of existing datasets and ours.}
\label{tab:bangla-hate-datasets}
\end{table}

\subsection{Existing Hate Speech Datasets}
There has been effort to develop datasets in the past. \citet{gupta2022macd} introduced a 150K-comment dataset for abusive speech detection in five Indic languages, while \citet{sharif2021nlp} studied offensive language detection in multilingual code-mixed text. These work establish important baselines for future research on code-mixed offensive text detection in Dravidian languages \cite{saumya-etal-2021-offensive, chakravarthi2022dravidiancodemix}. 

Some of the notable resources for hate and abusive content on Bangla include 10,178 tweets labeled as hate/offensive/normal \cite{das2022hate}, a 30K comments dataset with 10K hate speech examples \cite{romim2020hate}, 3K transliterated Bangla-English abusive comments \cite{sazzed2021abusive}, 50K offensive comments from online social networking \cite{romim-etal-2022-bd}, and 10K Bangla posts consisting of 5K actual and 5K Romanized Bengali tweets \cite{das-etal-2022-hate-speech}. Moreover, a multi-label transliterated Bangla hate speech dataset has been developed by \citet{haider2024banth} utilizing a translation-based LLM prompting approach.

Building on these efforts, Table~\ref{tab:bangla-hate-datasets} provides an overview of existing resources. Our contribution extends this landscape by introducing a larger dataset that not only supports multiple tasks but also incorporates a richer topical hierarchy, spanning 19 topics and 120 sub-topics.

\section{Dataset}
\label{sec:dataset}


\subsection{Data Collection}

\begin{figure}[t]
    \centering    
    \includegraphics[scale=0.25]{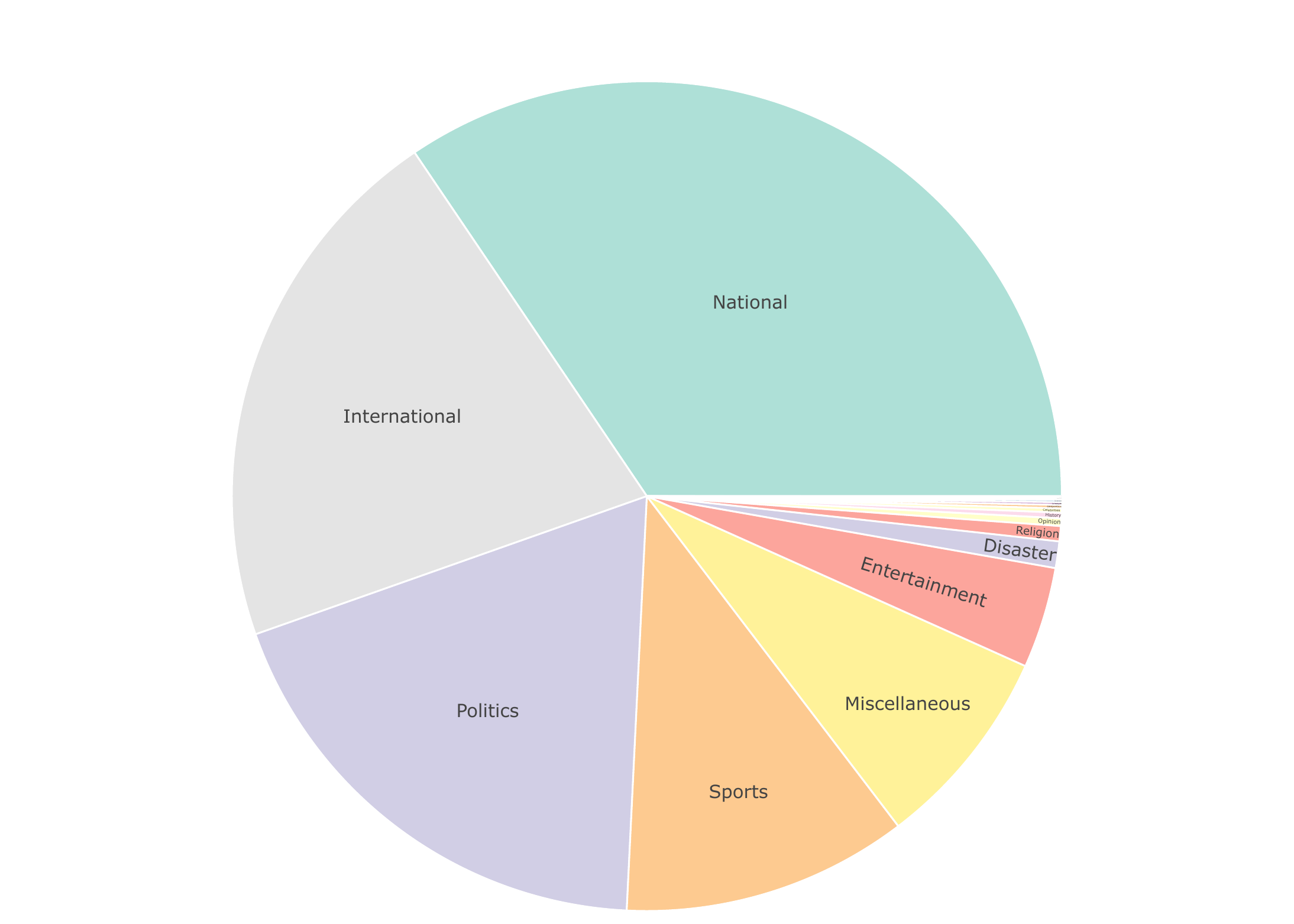}
    \vspace{-0.2cm}
    \caption{Distribution of the MultiHate dataset across different categories.}
    \label{fig:multihate_dataset_plot}
    \vspace{-0.4cm}
\end{figure}

We collected public comments from YouTube videos using the YouTube API\footnote{\url{https://developers.google.com/youtube/v3}}, primarily from Somoy TV, which is a popular Bangla News channel. The comments belong to 19 different categories, including \textit{Business, Celebrities, Disaster, Entertainment, Fashion, Geopolitics, Health, History, International, Lifestyle, Literature, Miscellaneous, National, Opinion, Politics, Religion, Science, Sports}, and \textit{Technology}, as well as 120 subcategories. In total, we collected approximately 55,000 comments associated with various Bangla news videos. We then removed all entries containing only emojis and URLs, as well as duplicate entries. Additionally, we excluded all Banglish comments (Bangla text written using the English alphabet) from the initial dataset. After applying these filtering and duplicate-removal steps, the dataset contained 50,746 entries. 
The category-wise data distribution is presented in Figure \ref{fig:multihate_dataset_plot}, with more than 90\% of the comments concentrated in five categories.

\subsection{Data Annotation}
\label{ssec:annotation}

\subsubsection{Annotation Guidelines}

We developed an annotation guideline to facilitate the annotation of data. Our annotation setup was a multitask   
annotation. Therefore, each instance could be assigned multiple labels to capture the overlapping and nuanced nature of the content, such as identifying the type of hate, the severity of hate, and the target of hate simultaneously. The guidelines provided clear definitions, decision criteria, and illustrative examples to ensure consistency across annotators. Below, we briefly discuss the annotation guidelines for each annotation task.

\begin{enumerate}[noitemsep,topsep=0pt,leftmargin=*,labelsep=.5em]

    \item \textbf{Type of Hate:} 
    The purpose of this task is to identify the type of hate from YouTube comments. The annotators classified whether the comments are \textit{Abusive, Sexism, Religious Hate, Political Hate, Profane,} or \textit{None} based on the criteria discussed in the Appendix \ref{sec:app_annotation_guideline}. Depending on the nature of the annotation, the annotator proceeds with different subsequent tasks. If a comment is marked as \textit{None}, annotators automatically assign the labels \textit{Little to None} for the severity of hate and \textit{None} for the target of hate; otherwise, they proceed with the regular annotation process.

    \item \textbf{Severity of Hate:} 
    This task aims to assess the degree of hate expressed in a given comment. Annotators evaluate whether the comment reflects \textit{Little to None, Mild,} or \textit{Severe} forms of hate, taking into account factors such as the intensity of derogatory expressions, the use of slurs, and the presence of threats or incitement to violence. The objective is to capture not only the presence of hateful content but also its strength and potential impact. In the annotation guidelines, clear criteria and illustrative examples are provided to ensure consistency and reliability among annotators. 

    \item \textbf{Target of Hate: }
    This task focuses on identifying the specific \textit{Individuals, Organizations, Communities, Society,} or \textit{None} that is the target of hateful expression. Annotators classify whether the hate is directed toward protected characteristics such as organizations, communities, or society, or if it is aimed at individuals without reference to group identity. In cases where no explicit target is present, annotators assign the label \textit{None}. The goal of this task is to capture the social dimension of hateful language, enabling analysis not only of the presence of hate but also of who or what is being targeted.  

\end{enumerate}

\subsubsection{Manual Annotation}
\label{sssec:manual_annotation}

The annotation team comprised 35 native Bangla-speaking undergraduate students, including both male and female annotators. The team was trained and supervised by expert annotators to ensure reliability and consistency in the labeling process. Each comment was independently annotated by three annotators, and periodic quality checks of randomly selected samples were conducted, followed by feedback sessions to maintain annotation standards. The final label for each comment was determined by majority agreement among the annotators. In instances of persistent disagreement, consensus meetings were organized to resolve discrepancies and establish the final annotation. 

\paragraph{Annotation Agreement} 
We evaluated the inter-annotator agreement (IAA) of the manual annotations using Fleiss’ Kappa coefficient ($\kappa$) to assess the reliability of the annotation process across all tasks. The obtained $\kappa$ scores were $0.71$, $0.84$, and $0.79$ for the type of hate, severity of hate, and target of hate tasks, respectively, indicating substantial to perfect agreement.\footnote{According to \citet{landis:koch:1977}, values of $\kappa$ between 0.61–0.80 represent substantial agreement, while values between 0.81–1.0 represent almost perfect agreement.} We also observed that an increase in the number of annotation classes introduces greater challenges for annotators, which is reflected in lower IAA scores for more fine-grained tasks. 

\subsection{Analysis and Statistics}
\label{ssec:data_stat}

We present the detailed class label distribution in Table \ref{tab:detailed_stats}. The table reports the frequency of labels across the three tasks, 
for the training, development, and test splits. This breakdown highlights the inherent class imbalance across tasks, with \textit{None} being the most frequent label, whereas categories such as \textit{Sexism} or \textit{Religious Hate} are underrepresented. Such skewed distributions demonstrate the challenge of building robust models capable of handling rare but socially significant cases of hate speech. Moreover, Table~\ref{tab:word_length_bin} presents the distribution of class labels across word length bins for the three tasks. The majority of samples in all splits fall within the $\leq$20, indicating that most instances of hate speech in Bangla are expressed concisely.

\begin{table*}[!ht]
\centering
\setlength{\tabcolsep}{3pt}
\scalebox{0.8}{
\begin{tabular}{llrrrrrrrrrr}
\toprule
\multicolumn{1}{c}{\multirow{2}{*}{\textbf{Split}}} & \multicolumn{1}{c}{\multirow{1}{*}{\textbf{Type of Hate}}} & \multicolumn{3}{c}{\textbf{Severity of Hate}} & \multicolumn{1}{l}{\multirow{2}{*}{\textbf{Total}}} & \multicolumn{5}{c}{\textbf{Target of Hate}} & \multicolumn{1}{l}{\multirow{2}{*}{\textbf{Total}}} \\ \cline{2-2} \cline{3-5}\cline{7-11}
\multicolumn{1}{c}{} & \multicolumn{1}{c}{\textbf{Class}} & \multicolumn{1}{l}{LN} & \multicolumn{1}{l}{Mild} & \multicolumn{1}{l}{Severe} & \multicolumn{1}{l}{} & \multicolumn{1}{l}{Comm.} & \multicolumn{1}{l}{Indiv.} & \multicolumn{1}{l}{Org.} & \multicolumn{1}{l}{Society} & \multicolumn{1}{l}{None} & \multicolumn{1}{l}{} \\ \midrule
\multirow{7}{*}{\textbf{Train}} & Abusive & 2,254 & 3,867 & 2,091 & \textbf{8,212} & 1,521 & 3,340 & 1,510 & 1,118 & 723 & \textbf{8,212} \\
 & Political Hate & 978 & 2,237 & 1,012 & \textbf{4,227} & 404 & 935 & 1,896 & 745 & 247 & \textbf{4,227} \\
 & Profane & 127 & 396 & 1,808 & \textbf{2,331} & 399 & 1,182 & 385 & 183 & 182 & \textbf{2,331} \\
 & Religious Hate & 150 & 301 & 225 & \textbf{676} & 283 & 119 & 51 & 147 & 76 & \textbf{676} \\
 & Sexism & 26 & 52 & 44 & \textbf{122} & 28 & 70 & 4 & 12 & 8 & \textbf{122} \\
 & None & 19,954 & -- & -- & \textbf{19,954} & -- & -- & -- & -- & 19,954 & \textbf{19,954} \\ \midrule
 & \textbf{Total} & \textbf{23,489} & \textbf{6,853} & \textbf{5,180} & \textbf{35,522} & \textbf{2,635} & \textbf{5,646} & \textbf{3,846} & \textbf{2,205} & \textbf{21,190} & \textbf{35,522} \\ \midrule
\multirow{7}{*}{\textbf{Dev}} & Abusive & 333 & 507 & 273 & \textbf{1,113} & 207 & 427 & 251 & 141 & 87 & \textbf{1,113} \\
 & Political Hate & 141 & 292 & 141 & \textbf{574} & 43 & 130 & 270 & 101 & 30 & \textbf{574} \\
 & Profane & 25 & 63 & 254 & \textbf{342} & 52 & 171 & 58 & 23 & 38 & \textbf{342} \\
 & Religious Hate & 16 & 36 & 26 & \textbf{78} & 28 & 18 & 4 & 17 & 11 & \textbf{78} \\
 & Sexism & 4 & 11 & 4 & \textbf{19} & 8 & 9 & 1 & 1 & \multicolumn{1}{l}{} & \textbf{19} \\
 & None & 2,898 & -- & -- & \textbf{2,898} & -- & -- & -- & -- & 2,898 & \textbf{2,898} \\ \midrule
 & \textbf{Total} & \textbf{3,417} & \textbf{402} & \textbf{698} & \textbf{5,024} & \textbf{338} & \textbf{755} & \textbf{584} & \textbf{283} & \textbf{3,064} & \textbf{5,024} \\ \midrule
\multirow{7}{*}{\textbf{Test}} & Abusive & 646 & 1,075 & 591 & \textbf{2,312} & 441 & 891 & 465 & 306 & 209 & \textbf{2,312} \\
 & Political Hate & 263 & 702 & 255 & \textbf{1,220} & 124 & 263 & 541 & 231 & 61 & \textbf{1,220} \\
 & Profane & 41 & 116 & 552 & \textbf{709} & 127 & 354 & 127 & 51 & 50 & \textbf{709} \\
 & Religious Hate & 29 & 93 & 57 & \textbf{179} & 62 & 47 & 17 & 34 & 19 & \textbf{179} \\
 & Sexism & 7 & 15 & 7 & \textbf{29} & 5 & 16 & 2 & 3 & 3 & \textbf{29} \\
 & None & 5,751 & -- & -- & \textbf{5,751} & -- & -- & -- & -- & 5,751 & \textbf{5,751} \\ \midrule
 & \textbf{Total} & \textbf{6,737} & \textbf{2,001} & \textbf{1,462} & \textbf{10,200} & \textbf{759} & \textbf{1,571} & \textbf{1,152} & \textbf{625} & \textbf{6,093} & \textbf{10,200} \\
 \bottomrule
\end{tabular}
}
\caption{Class label distribution of the dataset. LN: Little to None, Comm.: Community, Indiv.: Individual, Org.: Organization}
\label{tab:detailed_stats}
\end{table*}

\begin{figure}[!tbh]
    \centering    
    \includegraphics[scale=0.35]{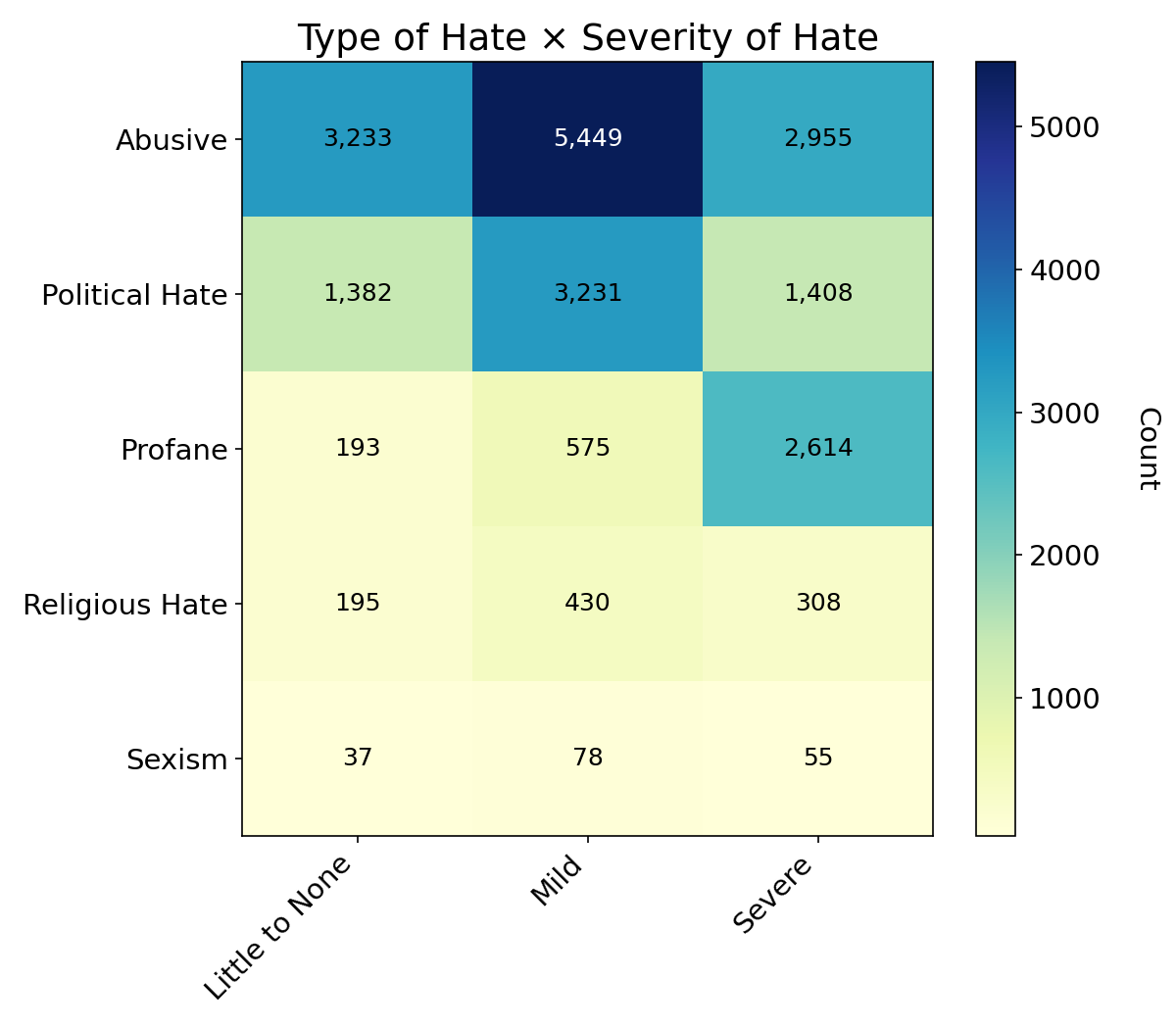}
    \vspace{-0.2cm}
    \caption{Heatmap demonstrating the relationship between \textit{type} of hate and \textit{severity}.}
    \label{fig:type_vs_severity_heatmap}
    \vspace{-0.4cm}
\end{figure}

\begin{figure}[!tbh]
    \centering    
    \includegraphics[scale=0.35]{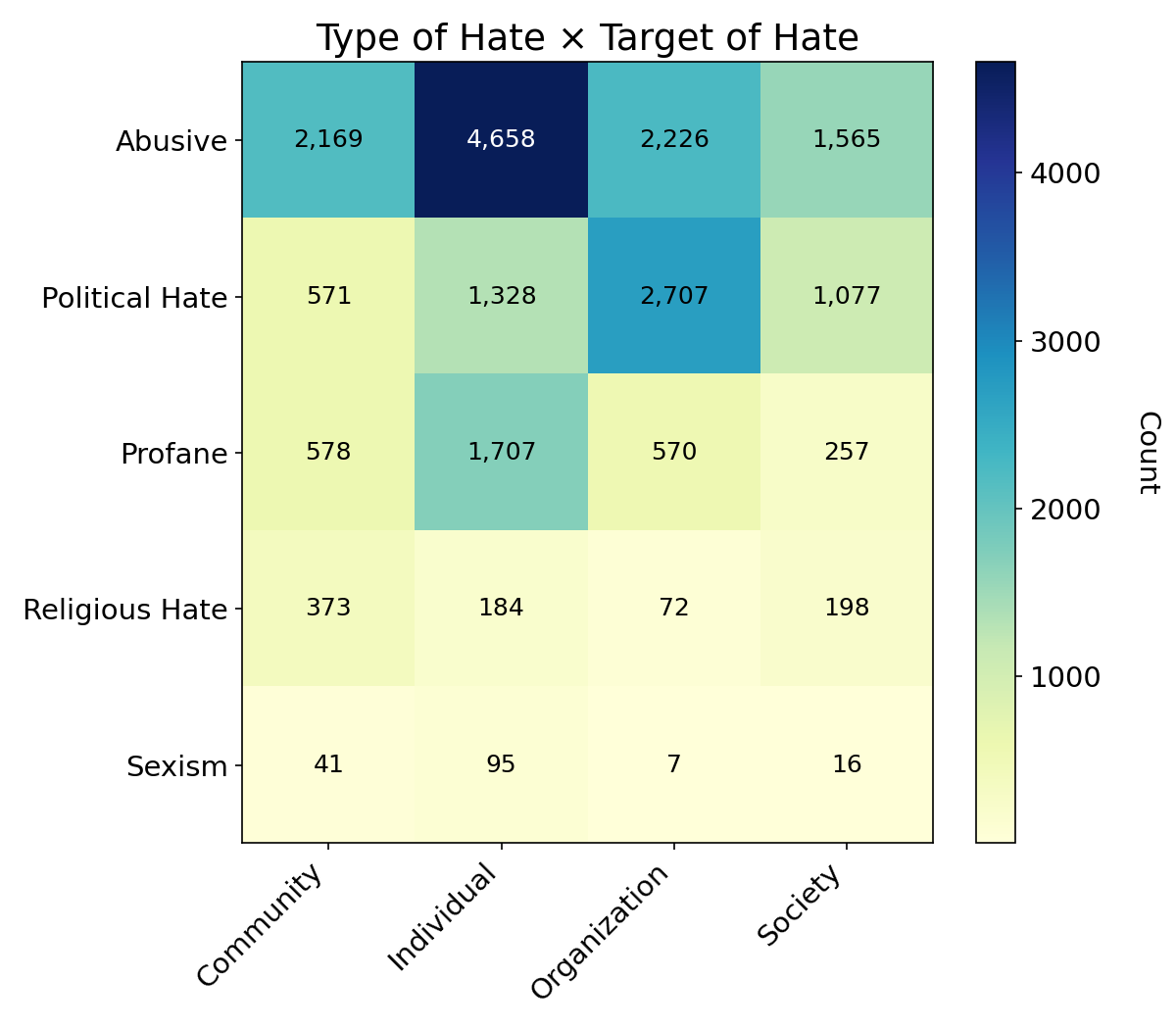}
    \vspace{-0.2cm}
    \caption{Heatmap demonstrating the relationship between \textit{type} of hate and \textit{target}.}
    \label{fig:type_vs_target_heatmap}
    \vspace{-0.4cm}
\end{figure}

In Figures \ref{fig:type_vs_severity_heatmap} and \ref{fig:type_vs_target_heatmap}, we present the relationship between hate type and severity, and between hate type and target, respectively. The \textit{None} category was excluded from the hate type for a concise visual representation.
Figure \ref{fig:type_vs_severity_heatmap} shows that \textit{Abusive} content is most prevalent, peaking at mild severity with a notable severe presence, while \textit{Political Hate} follows a smaller but similar trend. \textit{Profane} content stands out, concentrated in the severe category, suggesting profanity as a marker of high severity. \textit{Religious Hate} appears at low frequency across all severities, and \textit{Sexism} is rare overall. Mild severity emerges as the dominant category across types.  
Figure \ref{fig:type_vs_target_heatmap} highlights that individuals and organizations are the main targets, with abusive expressions disproportionately directed at individuals.

\subsection{Data Split}
\label{ssec:data_split}

The dataset was partitioned into training, development, and test sets, comprising 70\%, 10\%, and 20\% of data, respectively, for our experiments. We applied stratified sampling \cite{sechidis2011stratification} to ensure a balanced class label distribution across all splits. We provide the detailed data distribution across the splits in Table \ref{tab:data_distribution}. As shown, the dataset is highly imbalanced across all three annotation dimensions. For the \textit{type of hate} task, the majority of samples fall under the \textit{none} class, while categories such as \textit{sexism} and \textit{religious hate} are comparatively underrepresented. A similar pattern is observed in the \textit{severity of hate} task, where most comments are labeled as \textit{little to none}, followed by \textit{mild} and \textit{severe}. For the \textit{target of hate} task, the \textit{none} class again dominates, whereas labels such as \textit{society} and \textit{community} appear far less frequently. This imbalance highlights the inherent challenges of training reliable models on underrepresented classes and demonstrate the importance of stratification for fair evaluation. 

\begin{table}[!ht]
\scalebox{0.89}{
\begin{tabular}{lrrrr}
\toprule
\multicolumn{1}{c}{\textbf{Class}} &
  \multicolumn{1}{c}{\textbf{Train}} &
  \multicolumn{1}{c}{\textbf{Dev}} &
  \multicolumn{1}{c}{\textbf{Test}} &
  \multicolumn{1}{c}{\textbf{Total}} \\ \midrule
\multicolumn{5}{c}{\textbf{Type of Hate}}  \\ \midrule
Abusive        & 8,212           & 1,113          & 2,312           & \textbf{11,637} \\
Political Hate & 4,227           & 574           & 1,220           & \textbf{6,021}  \\
Profane        & 2,331           & 342           & 709            & \textbf{3,382}  \\
Religious Hate & 676            & 78            & 179            & \textbf{933}   \\
Sexism         & 122            & 19            & 29             & \textbf{170}   \\
None           & 19,954          & 2,898          & 5,751           & \textbf{28,603} \\ \midrule
\textbf{Total} & \textbf{35,522} & \textbf{5,024} & \textbf{10,200} & \textbf{50,746} \\ \midrule
\multicolumn{5}{c}{\textbf{Severity of Hate}} \\ \midrule
Severe         & 5,180           & 698           & 1,462           & \textbf{7340}  \\
Mild           & 6,853           & 909           & 2,001           & \textbf{9763}  \\
Litle to None  & 23,489          & 3,417          & 6,737           & \textbf{33,643} \\ \midrule
\textbf{Total} & \textbf{35,522} & \textbf{5,024} & \textbf{10,200} & \textbf{50,746} \\ \midrule
\multicolumn{5}{c}{\textbf{Target of Hate}}   \\ \midrule
Community      & 2,635           & 338           & 759            & \textbf{3,732}  \\
Individual     & 5,646           & 755           & 1,571           & \textbf{7,972}  \\
Organization   & 3,846           & 584           & 1,152           & \textbf{5,582}  \\
Society        & 2,205           & 283           & 625            & \textbf{3,113}  \\
None           & 21,190          & 3,064          & 6,093           & \textbf{30,347} \\ \midrule
\textbf{Total} & \textbf{35,522} & \textbf{5,024} & \textbf{10,200} & \textbf{50,746} \\ \bottomrule
\end{tabular}
}
\caption{Class label distribution across three tasks of the \textit{BanglaMultiHate} dataset.}
\label{tab:data_distribution}
\end{table}

\section{Methodology}
\label{sec:experiments}


\subsection{Models}
We experiment with classical model such as SVM, monolingual pretrained language model such as BanglaBERT \cite{bhattacharjee2021banglabert}, and large language models such as Llama-3.2-3B-Instruct\footnote{\href{https://huggingface.co/meta-llama/Llama-3.2-3B-Instruct}{Llama-3.2-3B-Instruct}} and Qwen3-4B-Instruct-2507\footnote{\href{https://huggingface.co/Qwen/Qwen3-4B-Instruct-2507}{Qwen3-4B-Instruct}}. We choose models from different model families to provide extensive evaluation with this dataset.

\noindent
\textbf{Baseline.} We used a majority-class baseline that always predicts the class with the highest frequency in the training data and a random approach. These methods have been widely used as a baseline technique in numerous prior studies (e.g., \cite{rosenthal2017semeval}).  

\noindent
\textbf{Classical models.} We employed SVM \cite{platt:98} with TF-IDF representation which has been extensively utilized in prior research and remains prevalent in low-resource production settings. Our setup employed 1–5 n-grams with TF-IDF weighting and a regularization parameter of $C=1$.



\noindent
\textbf{Pretrained Language Model (PLM).} Given that PLMs have demonstrated significant success in the past years and are also computationally reasonable choices for many downstream NLP tasks, we fine-tuned the monolingual BanglaBERT model \cite{wolf2020transformers}. Following the procedure of \citet{devlin2019bert}, we trained each model with default hyperparameters for 3 epochs. To mitigate training instability, we performed ten runs with different random seeds and selected the best model based on development set performance. All experiments were conducted independently for each task.



\noindent
\textbf{LLMs.} Recent advances in LLMs have received significant attention from researchers to evaluate the performance of these models, especially for low-resource languages in various downstream NLP tasks. We experiment with Llama-3.2-3B-Instruct~\cite{dubey2024llama} and Qwen3-4B-Instruct~\cite{yang2025qwen3}. We adopt a zero-shot learning setup for all models. To ensure reproducibility, we apply a consistent prompt, response format, output token limit, and decoding configuration (such as temperature set to 0) across models. The prompts were crafted using concise instructions, as detailed in Appendix \ref{sec:app_instruction_prompt}. We also demonstrate the efficacy of \textit{BanglaMultiHate} dataset by fine-tuning both Llama and Qwen models. We choose PEFT using LoRA \cite{hu2021lora} to reduce the computational cost. We trained the model in full precision (FP16) using the Adam optimizer. The learning rate was set to $2 \times 10^{-4}$, with LoRA parameters $\alpha=16$ and $r=64$. The maximum sequence length was fixed at $512$, and training was performed with a batch size of $8$. Fine-tuning was conducted for three epochs without additional hyperparameter tuning. Moreover, both zero-shot and fine-tuned approaches were studied in a multi-task setup due to computational resource constraints.

\subsection{Instructions Dataset}
We employed a template-based approach to generate diverse English instructions, obtaining 10 hate speech classification task templates per language from GPT-4.1 and Claude-3.5 Sonnet.\footnote{\href{https://www.anthropic.com/news/claude-3-5-sonnet}{claude-3-5-sonnet}} 
During fine-tuning and inference, one template was randomly selected and combined with the comment. We report examples of prompts in Appendix~\ref{sec:app_instruction_prompt}.


\subsection{Evaluation Measures}
Across all experimental settings, we evaluate performance using accuracy, micro-F1 score, as well as weighted precision and recall, with the weighted metrics chosen to account for class imbalance.

\section{Results and Discussion}
\label{sec:results}

In Table \ref{tab:results_full}, we present the performance of different models across all three tasks. 
Results are reported in terms of accuracy, precision, recall, and micro-F1.

\subsection{Comparison with Baselines} 
Across all three tasks, both the SVM and pretrained language models substantially outperform the majority and random baselines. Although the majority baseline achieves relatively high accuracy in the hate severity task, this is largely due to label imbalance. Zero-shot results consistently surpass the random baseline across all tasks, yet they still fall behind the majority baseline, demonstrating their limitations without task-specific adaptation. In contrast, PEFT with LoRA yields notable gains. In particular, fine-tuned Llama3 outperforms both baselines across all three tasks, demonstrating the effectiveness of fine-tuning the model. Qwen3 with LoRA shows modest improvements, performing slightly above the majority baseline.

\begin{table}[!ht]
\scalebox{0.9}{
\begin{tabular}{lrrrr}
\toprule
\textbf{Model} &
  \multicolumn{1}{l}{\textbf{Acc.}} &
  \multicolumn{1}{l}{\textbf{P.}} &
  \multicolumn{1}{l}{\textbf{R.}} &
  \multicolumn{1}{l}{\textbf{F1}}
  \\ \midrule
\multicolumn{5}{c}{\textbf{Type of Hate}}   \\ \midrule
Majority baseline & 0.564 & 0.318 & 0.564 & 0.564\\
Random baseline & 0.164 & 0.385 & 0.164 & 0.164 \\
SVM   & 0.609 & 0.574 & 0.609 & \textbf{0.609} \\
BanglaBERT        & 0.712 & 0.716 & 0.712 & \underline{\textbf{0.712}} \\ \midrule
\multicolumn{5}{c}{\textbf{Zero-Shot}} \\ \midrule
Llama3            & 0.275 & 0.619 & 0.275 & 0.275\\
Qwen3 & 0.520 & 0.542 & 0.520 & 0.520 \\ \midrule
\multicolumn{5}{c}{\textbf{Fine-tuned}}     \\ \midrule
Llama3            & 0.620 & 0.725 & 0.620 & \textbf{0.620} \\
Qwen3             & 0.595 & 0.453 & 0.595 & \textbf{0.595} \\ \midrule
\multicolumn{5}{c}{\textbf{Severity of Hate}} \\ \midrule
Majority baseline & 0.660 & 0.436 & 0.660 & 0.660 \\
Random baseline & 0.327 & 0.486 & 0.327 & 0.327 \\
SVM               & 0.672 & 0.607 & 0.672 & \textbf{0.672} \\
BanglaBERT        & 0.722 & 0.727 & 0.722 & \underline{\textbf{0.722}} \\ \midrule
\multicolumn{5}{c}{\textbf{Zero-Shot}} \\ \midrule
Llama3            & 0.508 & 0.729 & 0.508 & 0.508 \\
Qwen3             & 0.589 & 0.639 & 0.589 & 0.589 \\ \midrule
\multicolumn{5}{c}{\textbf{Fine-tuned}}  \\ \midrule
Llama3            & 0.685 & 0.682 & 0.685 & \textbf{0.685} \\
Qwen3             & 0.661 & 0.436 & 0.661 & \textbf{0.661} \\ \midrule
\multicolumn{5}{c}{\textbf{Target of Hate}}  \\ \midrule
Majority baseline & 0.597 & 0.357 & 0.597 & 0.597 \\
Random baseline & 0.204 & 0.404 & 0.204 & 0.204 \\
SVM               & 0.629 & 0.568 & 0.629 & \textbf{0.629} \\
BanglaBERT        & 0.715 & 0.716 & 0.715 & \underline{\textbf{0.715}} \\ \midrule
\multicolumn{5}{c}{\textbf{Zero-Shot}}  \\ \midrule
Llama3            & 0.340 & 0.465 & 0.340 & 0.340 \\
Qwen3             & 0.434 & 0.508 & 0.434 & 0.434 \\ \midrule
\multicolumn{5}{c}{\textbf{Fine-tuned}} \\ \midrule
Llama3            & 0.610 & 0.716 & 0.610 & \textbf{0.610} \\
Qwen3             & 0.598 & 0.470 & 0.598 & \textbf{0.598} \\
\bottomrule
\end{tabular}
}
\caption{Performance of different models on Bangla hate speech detection across three tasks. Acc.: Accuracy, P.: Precision, R.: Recall, F1: micro-F1 score. \textbf{Bold} indicates results that surpass both baseline methods for the respective task, while \underline{Underline} denotes the best overall performance across all three tasks.}
\label{tab:results_full}
\end{table}

\subsection{Classical Model and PLMs}
SV trained with lexical features such as n-grams and TF-IDF, provides consistent but modest improvements over both baselines. For instance, in \textit{type of hate} task, the SVM achieves 0.609 micro-F1 compared to 0.564 micro-F1 from the majority classifier. Similar improvements are seen in the \textit{target of hate} task. These results suggest that traditional supervised methods can exploit word-level patterns that naive baselines miss, making them moderately effective for detecting stereotypical hate expressions. However, the performance changes when used with narrow or context-dependent forms of hate speech, such as implicit derogatory references. This limitation stems from their reliance on surface-level features without deeper semantic understanding.

Leveraging pretraining on large-scale Bangla corpora, BanglaBERT consistently outperforms all models across all tasks. These gains highlight the importance of contextual embeddings from language-specific pretrained models, which enable the system to capture semantic nuances, idiomatic expressions, and subtle markers of abusive tone.

\subsection{Zero-shot Learning}
Across all three tasks Llama3 and Qwen3 show mixed performance. For \textit{type of hate}, Llama3 achieves micro-F1 score of 0.275, which is only marginally better than the random baseline (0.164), highlighting the difficulty of zero-shot classification in datasets with imbalanced labels. Qwen3 performs substantially better in the same task with an accuracy and F1 of 0.520, performing better than Llama3 and approaching the majority baseline (0.564), demonstrating that model size significantly influence zero-shot performance.

In the \textit{severity of hate} task, Llama3 and Qwen3 achieve micro-F1 score of 0.508 and 0.589, respectively. Both models perform better than the random baseline (0.327) but could not perform better than the majority baseline (0.660), suggesting that the inherent structure of the severity task, requires nuanced understanding of language intensity, limits the effectiveness of zero-shot approaches. For the \textit{target of hate}, the models achieve 0.340 (Llama3) and 0.434 (Qwen3) micro-F1 score, similarly perform better the random but below the majority baseline, indicating that identifying the \textit{target of hate} often requires explicit task-specific knowledge that zero-shot models may not fully possess. Overall, zero-shot learning provides a reasonable starting point, particularly for Qwen3, but generally falls short of majority baseline, emphasizing the need for task adaptation to achieve high performance.

\subsection{Fine-tuning LLMs}
Fine-tuning with LoRA significantly improves performance across all three tasks. For \textit{type of hate} task, Llama3 achieves micro-F1 score of 0.620, surpassing both zero-shot Llama3 (0.275) and Qwen3 (0.520), and also performs better than majority baseline. Fine-tuned Qwen3 achieves 0.595 micro-F1, slightly below Llama3, however, still demonstrating improvement over its zero-shot experiment. These results show that fine-tuning enables the models to better capture nuanced patterns. 

In the \textit{severity of hate} task, Llama3 achieves a micro-F1 score of 0.685, while Qwen3 obtains 0.661. Both models outperform the majority baseline (0.660), demonstrating the effectiveness of task-specific adaptation in assessing the intensity of hateful content. Similarly, in the \textit{target of hate} task, Llama3 reaches a micro-F1 score of 0.610, with Qwen3 achieving 0.598, marking a clear improvement over zero-shot performance. These results demonstrate that LoRA fine-tuning enables the models to capture subtle contextual cues that indicate the intended target of hate. Overall, LoRA fine-tuning effectively transforms pre-trained LLMs into task-aware classifiers, narrowing the performance gap with classical models like SVM and pretrained models such as BanglaBERT, while providing a scalable approach for hate speech detection in low-resource languages.

\subsection{Findings}
\noindent\textbf{Does language-specific pretraining improve Bangla hate-speech classification across tasks?}\
\textit{BanglaBERT} achieves the highest scores on all three tasks, indicating that pretraining on linguistically and culturally relevant Bangla data is most effective, especially for fine-grained distinctions such as severity and target.

\noindent\textbf{Are zero-shot LLMs sufficient, or is task-specific fine-tuning required?}\
Zero-shot approaches are insufficient for Bangla hate speech. Task-specific fine-tuning substantially improves LLMs performance; fine-tuned \textit{Llama3} is a promising alternative, whereas \textit{Qwen3} exhibits weaker gains, suggesting differences in pretraining data and alignment.

\noindent\textbf{How do fine-tuned LLMs compare to monolingual PLMs?}\
Fine-tuned LLMs performs reasonably, however, do not surpass \textit{BanglaBERT}. This shows that the continued importance of language-specific pretraining for reliable detection in low-resource settings.

\noindent\textbf{What dataset properties most affect evaluation?}\
Class imbalance inflates baseline performance (e.g., majority-class predictions, particularly for the severity task). Robust evaluation should report macro-F1 and per-class metrics and consider stratified splits.

\noindent\textbf{Does task-specific training help uniformly across tasks?}\
Task-specific training improves performance on all three tasks, with the largest practical benefits for nuanced categories (e.g., severity levels and target groups), where generic zero-shot models struggle.



\section{Conclusions and Future Work}
\label{sec:conclusions}

In this study, we present \multihate{}, a Bangla hate-speech dataset that is among the largest manually annotated corpora in a multi-task setting. The dataset comprises approximately 51K instances spanning 19 topics and 120 sub-topics. To demonstrate its utility, we conduct comprehensive experiments comparing classical approaches, pretrained language models (PLMs), and large language models (LLMs). Our findings underscore the importance of language-specific pretraining as well as task-specific fine-tuning for robust performance. As future work, we plan to extend the dataset with reasoning annotations to support task-level interpretability and explanation.

\section{Limitations}
This study has several limitations. First, our dataset is collected from YouTube comments, which contain examples that may be disturbing or offensive to readers. During the annotation process, annotators were explicitly cautioned about this content and provided with appropriate warnings. Second, from a modeling perspective, the dataset is highly imbalanced across classes, which may affect both training stability and performance evaluation. Addressing these issues, for example, through data augmentation, re-sampling strategies, or collecting additional underrepresented examples, remains an important direction for future work.

\section*{Ethics and Broader Impact}
Our dataset consists solely of comments and does not include any personally identifiable user information, thereby posing no direct privacy risks. Nonetheless, it is important to acknowledge that annotation is inherently subjective, which can introduce biases into the dataset. To mitigate this, we designed a clear annotation schema and provided detailed guidelines to annotators, aiming to ensure greater consistency and reliability. However, we encourage researchers and practitioners to remain careful of these limitations when using the dataset for model development or further studies.

Despite these issues, the dataset holds significant potential for positive societal impact. Models trained on \multihate{} can support social media platforms in identifying and moderating harmful content, thereby contributing to healthier online discourse.

\bibliography{bibliography/main}

\appendix

\section*{Appendix}

\section{Detailed Annotation Guideline}
\label{sec:app_annotation_guideline}

\subsection{Definitions}
The primary goal of this annotation task is to categorize YouTube comments in the Bangla language into specific categories based on the nature and severity of hate speech they contain, as well as identifying the target of such speech. 

\paragraph{Type of Hate Categorization} This annotation task involves having annotators annotate Bangla text samples according to the type of hate expressed. Each text is categorized into one of six classes: \textit{Abusive}, \textit{Sexism}, \textit{Religious Hate}, \textit{Political Hate}, \textit{Profane}, or \textit{None}. The goal is to capture the specific nature of hateful content, enabling models to distinguish between different forms of hate speech and non-hateful content.
\begin{itemize}
    \item \textbf{Abusive:} Comments that are directly insulting, intending to belittle or harm someone's dignity. For example, 
    (\textit{English: You are completely useless}). This comment degrades someone by calling them utterly useless.
    \item \textbf{Political Hate:} Comments that display hostility towards political beliefs, parties, or figures. For example, 
    (\textit{English: All political leaders are thieves}).
    \item \textbf{Profane:} Comments that use swear words or vulgar language, intended to shock or offend without targeting anyone specifically. For example, 
    (\textit{English: Son of a pig, you got guts}). This comment uses profanity to express frustration.
    \item \textbf{Religious Hate:} Comments targeting individuals or groups based on their religion or religious beliefs. For example, 
    (\textit{English: All people in this religion are bad}). This comment generalizes a whole religion as bad.

    \item \textbf{Sexism:} Comments that discriminate or belittle someone based on their gender, often reflecting stereotypes. For example, 
    (\textit{Women should cook only}). This comment reinforces the stereotype expression.
    \item \textbf{None:} Comments that do not exhibit hate or negativity, including neutral or positive comments. For example, 
    (\textit{English: I saw the movie today}). This comment do not reflect any hate content.
    
\end{itemize}

\paragraph{Severity of Hate} This annotation task involves having annotators label Bangla text samples according to the intensity of hateful content expressed. The severity is categorized into three levels: \textit{Severe}, \textit{Mild}, and \textit{Little to None}. This classification scheme is designed to capture the varying degrees of harmfulness in hate speech.
\begin{itemize}
    \item \textbf{Severe:} Comments that contain threats, extreme prejudice, or are highly offensive. For example, 
    (\textit{English: Get out of the house, I'll take care of a beast like you}).
    \item \textbf{Mild:} Comments that are derogatory or mildly offensive but do not contain threats. For example, 
    (\textit{English: You are a traitor, you will burn in hell as soon as you die}). 
    \item \textbf{Little to None:} Comments that are slightly negative, ambiguous, or completely neutral or positive. For example, 
    (\textit{English: Hanif transport should be stopped}).
\end{itemize}

\paragraph{Target of Hate} This annotation task involves having annotators label Bangla text samples based on the intended target of the hateful expression. The labels are divided into five categories: \textit{Community}, \textit{Individual}, \textit{Organization}, \textit{Society}, and \textit{None}. This categorization aims to capture whether the hate speech is directed at a specific person, a collective group, broader societal structures, or institutions, while also accounting for instances where no explicit target is present.
\begin{itemize}
    \item \textbf{Community:} Comments against a specific racial, ethnic, gender, or religious group. For example, 
    (\textit{English: People from this community are not trustworthy}). 
    \item \textbf{Individual:} Comments targeting a specific person, either by name or implication. For example, 
    (\textit{English: Sheikh Hasina you are thief, the thief's mother has a loud voice}).
    \item \textbf{Organization:} Comments aimed at specific companies, governmental bodies, or any formal group. For example, 
    (\textit{English: Somoy television seems to be government's}).
    \item \textbf{Society:} Comments that critique societal norms, values, or general community practices. For example, 
    (\textit{English: Italians do not get food, they will increase birth rate again}). 
    \item \textbf{None:} Comments that are ambiguous, or completely neutral or positive. For example, 
    (\textit{English: It should be like this: it's hard to bring politics into any game}).
\end{itemize}

\begin{table*}[!ht]
\centering
\begin{tabular}{lllrrrrrr}
\toprule
\multirow{2}{*}{\textbf{Split}} & \multirow{2}{*}{\textbf{Task}} & \multirow{2}{*}{\textbf{Label}} & \multicolumn{6}{c}{\textbf{Word Length Bins}} \\ \cline{4-9}
 &  &  & \multicolumn{1}{l}{\textless{}=10} & 11-20 & \multicolumn{1}{l}{21-30} & \multicolumn{1}{l}{31-40} & \multicolumn{1}{l}{41-50} & \multicolumn{1}{l}{51+} \\ \midrule
\multirow{14}{*}{\textbf{Train}} & \multirow{6}{*}{Type of Hate} & Abusive & 4374 & 2438 & 801 & 311 & 115 & 173 \\
 &  & Political Hate & 1614 & 1415 & 625 & 259 & 139 & 175 \\
 &  & Profane & 1329 & 624 & 214 & 72 & 45 & 47 \\
 &  & Religious Hate & 281 & 233 & 81 & 42 & 19 & 20 \\
 &  & Sexism & 57 & 36 & 14 & 9 & 0 & 6 \\
 &  & None & 12624 & 4814 & 1353 & 508 & 265 & 390 \\ \cline{2-9}
 & \multirow{3}{*}{Severity of Hate} & Little to None & 14433 & 5840 & 1699 & 683 & 336 & 498 \\
 &  & Mild & 3207 & 2176 & 825 & 307 & 153 & 185 \\
 &  & Severe & 2639 & 1544 & 564 & 211 & 94 & 128 \\ \cline{2-9}
 & \multirow{5}{*}{Target of Hate} & Community & 1131 & 888 & 336 & 134 & 62 & 84 \\
 &  & Individual & 3146 & 1552 & 539 & 202 & 89 & 118 \\
 &  & Organization & 1755 & 1259 & 470 & 175 & 86 & 101 \\
 &  & Society & 951 & 694 & 293 & 124 & 58 & 85 \\
 &  & None & 13296 & 5167 & 1450 & 566 & 288 & 423 \\ \midrule
\multirow{14}{*}{\textbf{Dev}} & \multirow{6}{*}{Type of Hate} & Abusive & 599 & 315 & 121 & 37 & 16 & 25 \\
 &  & Political Hate & 212 & 197 & 92 & 33 & 16 & 24 \\
 &  & Profane & 190 & 104 & 26 & 6 & 7 & 9 \\
 &  & Religious Hate & 27 & 35 & 9 & 0 & 4 & 3 \\
 &  & Sexism & 11 & 7 & 0 & 1 & 0 & 0 \\
 &  & None & 1809 & 717 & 185 & 87 & 42 & 58 \\ \cline{2-9}
 & \multirow{3}{*}{Severity of Hate} & Little to None & 2079 & 864 & 239 & 106 & 54 & 75 \\
 &  & Mild & 413 & 300 & 118 & 30 & 23 & 25 \\
 &  & Severe & 356 & 211 & 76 & 28 & 8 & 19 \\ \cline{2-9}
 & \multirow{5}{*}{Target of Hate} & Community & 141 & 111 & 49 & 18 & 6 & 13 \\
 &  & Individual & 429 & 203 & 76 & 19 & 12 & 16 \\
 &  & Organization & 261 & 196 & 78 & 23 & 11 & 15 \\
 &  & Society & 124 & 92 & 33 & 13 & 11 & 10 \\
 &  & None & 1893 & 773 & 197 & 91 & 45 & 65 \\ \midrule
\multirow{14}{*}{\textbf{Test}} & \multirow{6}{*}{Type of Hate} & Abusive & 1224 & 711 & 228 & 68 & 40 & 41 \\
 &  & Political Hate & 434 & 431 & 183 & 91 & 23 & 58 \\
 &  & Profane & 371 & 207 & 79 & 32 & 9 & 11 \\
 &  & Religious Hate & 73 & 59 & 28 & 12 & 4 & 3 \\
 &  & Sexism & 17 & 10 & 1 & 1 & 0 & 0 \\
 &  & None & 3645 & 1390 & 373 & 159 & 59 & 125 \\ \cline{2-9}
 & \multirow{3}{*}{Severity of Hate} & Little to None & 4153 & 1678 & 484 & 204 & 73 & 145 \\
 &  & Mild & 885 & 680 & 243 & 96 & 38 & 59 \\
 &  & Severe & 726 & 450 & 165 & 63 & 24 & 34 \\ \cline{2-9}
 & \multirow{5}{*}{Target of Hate} & Community & 329 & 242 & 106 & 46 & 16 & 20 \\
 &  & Individual & 861 & 468 & 136 & 54 & 25 & 27 \\
 &  & Organization & 494 & 399 & 142 & 59 & 20 & 38 \\
 &  & Society & 260 & 206 & 91 & 36 & 11 & 21 \\
 &  & None & 3820 & 1493 & 417 & 168 & 63 & 132 \\ \bottomrule
\end{tabular}

\caption{Detailed class label distribution with word length bin count.}
\label{tab:word_length_bin}
\end{table*}

\section{Prompts}
\label{sec:app_instruction_prompt}

We provide the instructions used to generate prompts for all three tasks in Listings~\ref{lst:prompt_generating_instruction_type}, \ref{lst:prompt_generating_instruction_severity}, and \ref{lst:prompt_generating_instruction_target}. Additionally, the prompts employed for zero-shot learning, fine-tuning, and inference are presented in Listing~\ref{lst:prompt_fine_tune_inference}.

\begin{lstlisting}[language=TeX,caption={Prompt for generating instructions for Type of Hate task.},label={lst:prompt_generating_instruction_type}]
We are creating an English instruction-following dataset for Type of Hate hate speech detection.
Read the given text carefully and choose the most appropriate label for the task from the label lists.
For the 'Type of Hate' task, the labels are 'Abusive', 'Sexism', 'Religious Hate', 'Political Hate', 'Profane', and 'None'. Select only one correct label for each task based on the information provided in the text and return your response in the following json format.
{"type_of_hate": "Abusive"}
    
Write 10 very diverse and concise English instructions. Only return the instructions without additional text. Do not generate additional text.
    
Return the instructions in a list format as follows.
['sent1', 'sent2']
\end{lstlisting}

\begin{lstlisting}[language=TeX,caption={Prompt for generating instructions for Severity of Hate task.},label={lst:prompt_generating_instruction_severity}]
We are creating an English instruction-following dataset for Hate Severity hate speech detection. Here is an example instruction:
Read the given text carefully and choose the most appropriate label for the task from the label lists.
For the 'Hate Severity' task, the labels are 'Little to None', 'Mild', and 'Severe'. Select only one correct label for each task based on the information provided in the text and return your response in the following json format.
{"severity_of_hate": "Mild"}
    
Write 10 very diverse and concise English instructions. Only return the instructions without additional text. Do not generate additional text.
    
Return the instructions in a list format as follows.
['sent1', 'sent2']
\end{lstlisting}

\begin{lstlisting}[language=TeX,caption={Prompt for generating instructions for Target of Hate task.},label={lst:prompt_generating_instruction_target}]
We are creating an English instruction-following dataset for the Target of Hate task of hate speech detection. Here is an example instruction: 
Read the given text carefully and choose the most appropriate label for the task from the label lists."+
For the 'Target of Hate' task, the labels are 'Individuals', 'Organizations', 'Communities', 'Society', and 'None'. Select only one correct label for the task based on the information provided in the text and return your response in the following json format.
{"type_of_hate": "Society"}

Write 10 very diverse and concise English instructions. Only return the instructions without additional text. Do not generate additional text.
    
Return the instructions in a list format as follows.
['sent1', 'sent2']
\end{lstlisting}


\begin{lstlisting}[language=TeX,caption={Sample Prompt for zero-shot learning, model fine-tuning, and inference.},label={lst:prompt_fine_tune_inference}]

You are a Bangla AI assistant specialized in the hate speech detection task. Your task is to identify the correct label for the task.


Read the text and assign the correct labels for the type of hate, severity of hate, and target of hate. Return only the answer without any explanation, justification, or additional text.
For the 'Type of Hate' task, the labels are 'Abusive', 'Sexism', 'Religious Hate', 'Political Hate', 'Profane', and 'None'. 
For the 'Severity of Hate' task, the labels are 'Little to None', 'Mild', and 'Severe'. 
And for the 'Target of Hate' task, the labels are 'Individuals', 'Organizations', 'Communities', 'Society', and 'None'. 
Select only one correct label for each task based on the information provided in the text and return your response in the following JSON format. 

{
    "type_of_hate": "Abusive", 
    "severity_of_hate": "Mild", 
    "target_of_hate": "Society"
}

If you select the 'None' label for the 'Type of Hate' task, the labels for the 'Severity of Hate' and 'Target of Hate' tasks would be 'Little to None' and 'None'.

\end{lstlisting}

\section{Data Release}
\label{apndix:release}
The \textit{BanglaMultiHate} dataset 
will be released under the CC BY-NC-SA 4.0 -- Creative Commons Attribution 4.0 International License: \url{https://creativecommons.org/licenses/by-nc-sa/4.0/}.

\end{document}